\documentclass[10pt,twocolumn,letterpaper]{article}

\usepackage{iccv}
\usepackage{times}
\usepackage{epsfig}
\usepackage{graphicx}
\usepackage{amsmath}
\usepackage{amssymb}

\usepackage{epsfig}
\usepackage{graphicx}
\usepackage{amssymb}
\usepackage{multirow}
\usepackage{adjustbox}
\usepackage{rotating}
\usepackage{color}
\usepackage{authblk}
\usepackage[accsupp]{axessibility}  

\usepackage[breaklinks=true,bookmarks=false]{hyperref}

\iccvfinalcopy 

\def\iccvPaperID{****} 

\ificcvfinal\fi

\begin{document}

\title{Nuisance-Label Supervision: Robustness Improvement by Free Labels}



\author[1]{Xinyue Wei}
\author[2]{Weichao Qiu}
\author[2]{Yi Zhang}
\author[2]{Zihao Xiao}
\author[2]{Alan Yuille}
\affil[1]{University of California San Diego}
\affil[2]{Johns Hopkins University}


\maketitle
\ificcvfinal\thispagestyle{empty}\fi

\begin{abstract}
   In this paper, we present a Nuisance-label Supervision (NLS) module, which can make models more robust to nuisance factor variations. Nuisance factors are those irrelevant to a task, and an ideal model should be invariant to them. For example, an activity recognition model should perform consistently regardless of the change of clothes and background. But our experiments show existing models are far from this capability. So we explicitly supervise a model with nuisance labels to make extracted features less dependent on nuisance factors. Although the values of nuisance factors are rarely annotated, we demonstrate that besides existing annotations, nuisance labels can be acquired freely from data augmentation and synthetic data. Experiments show consistent improvement in robustness towards image corruption and appearance change in action recognition.
\end{abstract}

\section{Introduction}

Model robustness is an important topic because it decides on whether a model can work well in actual applications. Many factors can affect model robustness, and \textbf{spurious correlation} is one of them. Spurious correlation refers to using inaccurate information to make predictions. For example, \textit{playing football} and \textit{grass} are always bundled up in action datasets; if a model uses grass to predict the task label \textit{playing football}, it is relying on the spurious correlation between \textit{grass} and the task label, and when people play football in other scenes, it fails. For a certain task, we call factors related to the task as \textbf{essential factors} and others as \textbf{nuisance factors}. When training a model, the goal is to accurately make predictions depending on essential factors, while invariant to the change of nuisance factors. Therefore, we should prevent models from relying on spurious correlation between task labels and nuisance factors.

Increasing data diversity is a common way to solve spurious correlation however it suffers from problems. The strong correlations are hard to break, e.g. the football and grass pair commonly appear in most action datasets. Simply adding to training data diversity is not sufficient for removing the correlation between nuisance factors and task labels. Recent works illustrate this problem that ImageNet-trained models are still biased towards textures~\cite{geirhos2018imagenet}, action recognition models trained on UCF101 or Kinetics are sensitive to viewpoint and human appearance changes~\cite{lyu2020identity}. Therefore, exploration about proactively disentangling nuisance factors and task label is worth thinking about.

\begin{figure}[t]
\begin{center}
   \includegraphics[width=\linewidth]{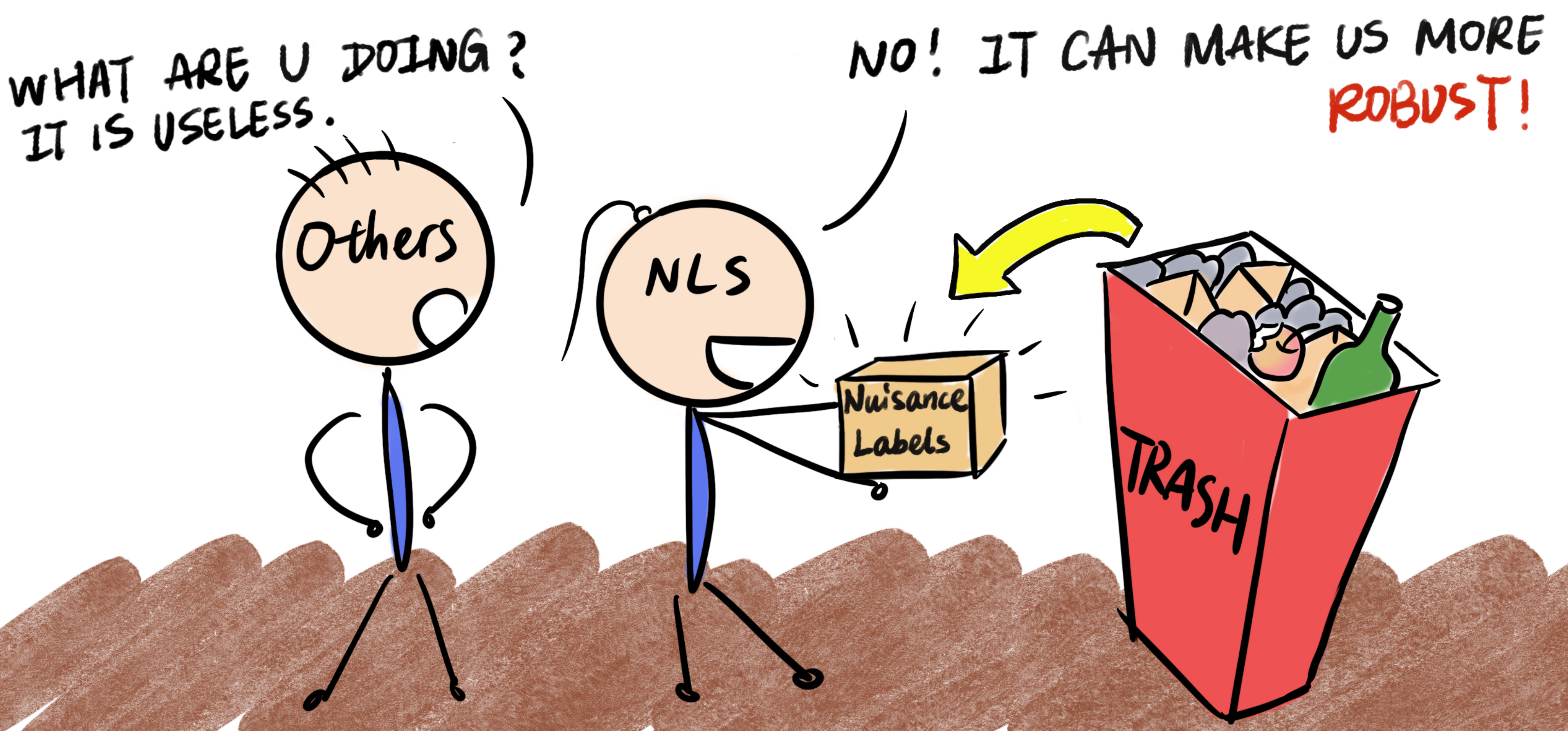}
\end{center}
    \vspace{-0.1in}
   \caption{Simple illustration of the core concept. Our proposed method Nuisance-label Supervision (NLS) module makes use of \textit{nuisance labels}, which is usually threw out by other methods, to improve model robustness.}
   \vspace{-0.1in}
\label{fig:teaser}
\end{figure}

We propose to explicitly use nuisance labels to train a feature representation robust to nuisance factor perturbations. We design a Nuisance-label Supervision (NLS) module, which utilizes nuisance labels as extra supervision signals and makes the whole network training in an adversary way. We first prove the core idea on MNIST-C which is widely used to study image corruption, then we demonstrate the method effectiveness on UCF101, and NTU RGB-D, which are both action datasets; the former is a popular dataset, but models trained on it usually fail to generalize to other datasets; the latter is a controlled dataset with multiple available nuisance labels. By using NLS, our trained model will be insensitive to viewpoint or human appearance changes.

Though NLS is simple and intuitive, collecting enough training data with nuisance labels is a non-trivial issue. We answer this challenging question by giving out three practical and low-cost ways of acquiring nuisance labels: 1) image processing parameters when doing data augmentation; 2) metadata in the data collection procedure; 3) rendering parameters from synthetic data generation. These three types of data are free and often ignored; Fig.~\ref{fig:teaser} vividly illustrates the idea of making waste profitable. We describe in detail how we get nuisance labels from these three sources in Sec.~\ref{section:sim_gen} and demonstrate the effectiveness in the experiment section. Previously, extra labels other than the task label have been proved effective in intermediate supervision~\cite{li2017deep} and multi-task learning~\cite{zamir2018taskonomy}. Compared with them, nuisance labels in our method are easy to collect because they all come from the processing steps of data generation.

Our module is generic and easy to use. It is not restricted to tasks or models and can be attached to any existing models, which makes our method widely applicable. The improvement by NLS is totally for free because the nuisance labels come from the data generation process, which costs no time to annotate. Besides, NLS only works during the training time and adds no extra calculation to inference.


In addition to NLS, we explore using nuisance labels for understanding model robustness. We diagnose model dependency on a certain nuisance factor and analyze its relationship with robustness. The results show the more a model relies on nuisance factors, the worse it performs on test set with different nuisance factor values.

The contribution can be summarized as follows: 
\begin{itemize}
    \item We propose a Nuisance-label Supervision (NLS) module to improve model robustness by breaking spurious correlation. The module is generic and can be applied to different tasks and models. 
    \item We propose three practical and low-cost ways of acquiring nuisance labels. Besides existing annotations, parameters from data augmentation and synthetic data can also be used for generating nuisance labels.
    \item Extensive experiments demonstrate the effectiveness of improving model robustness to image corruption and appearance changes. Nuisance labels are used to further understand over-fitting and model robustness improvement.
\end{itemize}
\section{Related Work}

\textbf{Robustness against irrelevant perturbations}. Recent works attempt to solve generalization failure by data augmentation and bias mitigation. Cubuk and Zoph~\cite{cubuk2019autoaugment} proposed AutoAugment for searching improved data augmentation policies; Hendrycks and Mu~\cite{hendrycks2019augmix} proposed image processing technique AugMix, Rusak and Schott~\cite{rusak2020increasing} introduced adding Gaussian noise to training data (GNT) and novel adversarial noise generator (ANT) to overcome unseen corruptions. These approaches focus on increasing data diversity while not making use of intermediate labels. Singh \textit{et al.}~\cite{singh2020don} used CAM~\cite{zhou2016learning} and feature-slitting methods to decorrelate category and its co-occurring context but only applicable to fixed category pairs. Approaches~\cite{alvi2018turning,kim2019learning} utilized extra labels to mitigate bias in feature representations, such as removing gender bias from age classification. While we focus on multiple factors, especially those rarely explored, and make use of parameters in synthetic data and data augmentation, which are cheaper labels compared with other work.

\textbf{Action Recognition}.
Action recognition model robustness to nuisance factors has gradually received more attention in recent years and is found to be fragile. Lyu \textit{et al.}~\cite{lyu2020identity} showed TSN~\cite{wang2016temporal} and I3D~\cite{carreira2017quo} make predictions relying on irrelevant information, such as related objects. Li \textit{et al.}~\cite{li2018resound} revealed that current action datasets are biased towards the scene, objects or people. Thus our work focuses on RGB-based models and tries to increase their robustness even learning from a biased dataset. Choi \textit{et al.}~\cite{choi2019can} introduce two adversarial losses to remove background effect while our work focus on multiple nuisance factors, including both 2D image corruptions and 3D factors e.g. viewpoint.

\textbf{Usage of extra labels}.
Intermediate supervision and multi-task focus on the usage of extra labels. Multi-task learning focuses on related tasks. e.g. pose estimation and action recognition~\cite{gkioxari2014r}, surface normals and depth~\cite{ren2018cross}, \textit{etc}. While we use labels not specifically annotated for any tasks, but parameter labels with less semantic meanings. DSN~\cite{lee2015deeply} is the first to propose the concept of deep supervision where the final label is used for supervising intermediate layers in a network. Furthermore, Li \textit{et al.}~\cite{li2017deep} utilized intermediate shape concepts acquired from rendered data to supervise CNN hidden layers and achieved synthetic to real generalization. Whereas nuisance labels used in our method is different from the final label or intermediate concepts, which are easy-acquired but often ignored.

\textbf{Synthetic data in vision tasks}. Synthetic data is widely used as augmentation and shows merits on vision tasks. The domain gap between synthetic and real is always a problem to solve and the various intermediate information during rendering is rarely utilized. Previously, domain randomization~\cite{tobin2017domain} is a popular way for bridging the real-sim gap. On the one hand, our proposed NLS helps reduce the domain gap and offers a good way for using extra information. On the other hand, synthetic data serves as a source of nuisance labels, providing a larger application space for our method, including action recognition~\cite{de2017procedural,zhang2019rsa,varol2019synthetic}, pose estimation~\cite{chen2016synthesizing,rogez2016mocap,rogez2018image}, segmentation~\cite{varol2017learning,Richter_2016_ECCV,ros2016synthia}, etc.

\section{Method}

\begin{figure*}[t]
\begin{center}
   \includegraphics[width=\linewidth]{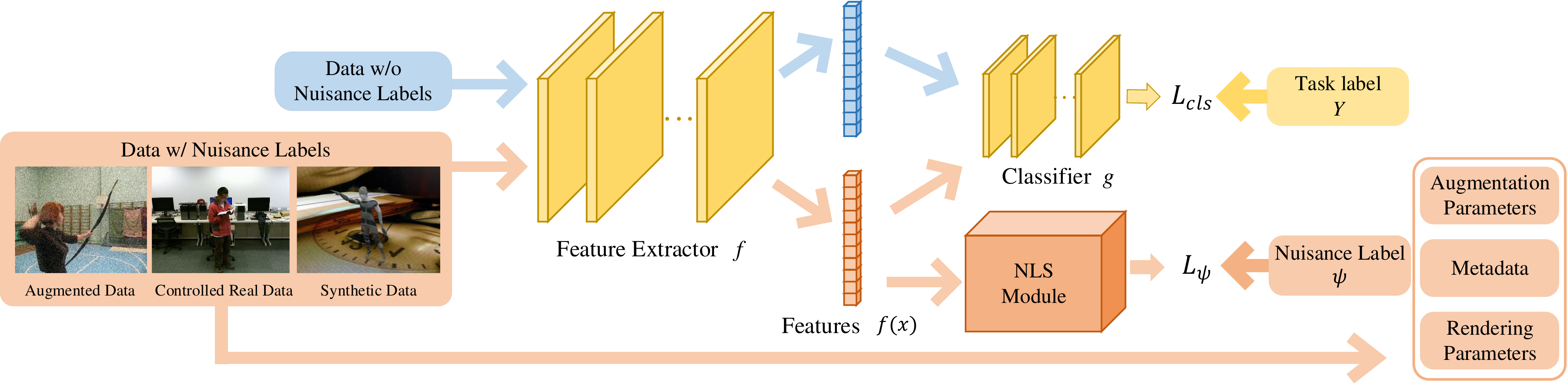}
\end{center}
    \vspace{-0.1in}
   \caption{The pipeline of the proposed method. The blue arrow represents data w/o nuisance labels while the red arrow represents data w/ nuisance labels. Both inputs are put through feature extractor $f$ and classifier $g$ as yellow arrow shows. Additionally, data with nuisance labels is put into NLS module for training feature representations. (Better view in color)}
   \vspace{-0.1in}
\label{fig:pipeline}
\end{figure*}

\begin{figure}[t]
\begin{center}
   \includegraphics[width=\linewidth]{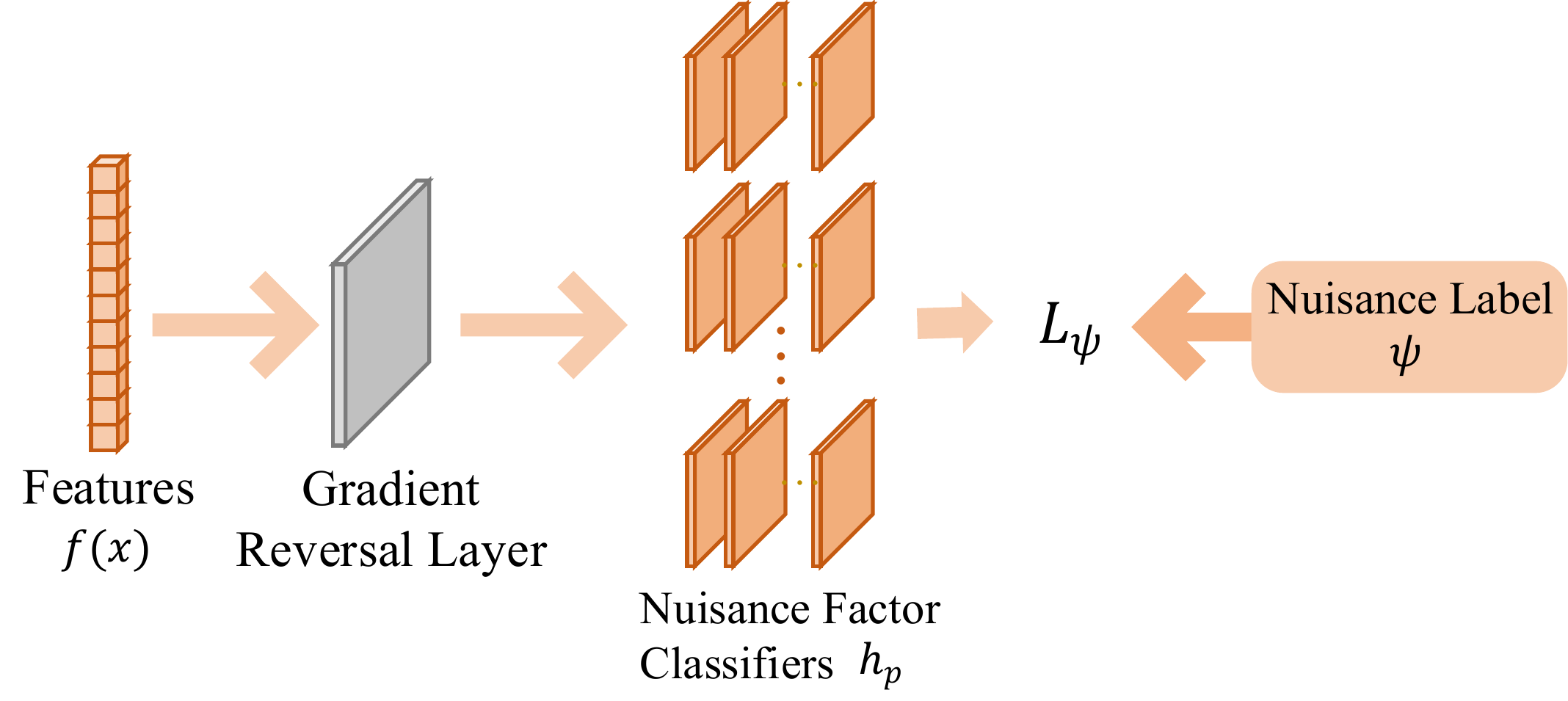}
\end{center}
    \vspace{-0.1in}
   \caption{The architecture of Nuisance-label Supervision Module. The module consists of gradient reversal layer and nuisance factor classifiers. The module input is extracted features and  nuisance labels. And the module is supervised by the nuisance adversarial loss $L_{\psi}$.}
   \vspace{-0.1in}
\label{fig:msm}
\end{figure}

This section can be roughly divided into four parts. First, we formulate the problem of using nuisance labels to improve model robustness (Sec.~\ref{section:prob_stat}). Second, we illustrate the structure and adversarial training procedure of our proposed NLS module (Sec.~\ref{section:msm}). Third, we show the generation of nuisance labels and synthetic data (Sec.~\ref{section:sim_gen}). Finally, we define \textit{Dependency Degree on Nuisance Factor} and study it with nuisance labels (Sec.~\ref{section:define_dep_degree}).

\subsection{Problem Statement}\label{section:prob_stat}
The objective of our work is to weaken the influence of nuisance factor perturbations by making full use of nuisance labels. We demonstrate our method on action recognition tasks because the action model is easily affected by various factors.

We aim to train a model robust to nuisance factor perturbations, especially when training and testing data have different nuisance factor values. We suppose the process of making prediction $y$ is written as:

\begin{equation}
    y = F(x) = F(D(\phi, \psi))
\end{equation}
where $F$ denotes the model, $y$ denotes model prediction, the factors for generating input $x \in X$ can be divided into essential factors $\phi \in \Phi$ and nuisance factors $\psi \in \Psi$, $D$ represents data generator. Thus an ideal model robust to nuisance factors should meet the following characteristics:

\begin{equation}
    \forall \psi_i, \psi_j \in \Psi, F(D(\phi, \psi_i)) = F(D(\phi, \psi_j))
\end{equation}
where no matter what value nuisance factors take, as long as essential factors do not change, final prediction should stay the same.

In order to better illustrate the problem, we give the definition of important terms. Examples specific to action recognition are given after the definition of each term.

\noindent\textbf{Nuisance Factors ($\Psi$)}: factors irrelevant to final prediction. The values of nuisance factors are nuisance labels and can be acquired from the data generation procedure. In this work, we divide the nuisance labels into three types according to their resources: 1) image processing parameters; 2) metadata in the data collection procedure; 3) rendering parameters from synthetic data generation. The details are shown in Sec.~\ref{section:sim_gen}

\noindent\textbf{Essential Factors ($\Phi$)}: factors decisive to final prediction. For action recognition, $\Phi$ is human motion, which is directly relevant to action definition.

\noindent\textbf{Task Label ($Y$)}: label for the main task. Task label does not necessarily equal essential factor labels, it is decided by essential factor values. Task label is unique in a certain task while essential factors may be more than one. For example, the essential factor in action recognition is human motion, and the task label is an action class.

To achieve the goal, we want the extracted features to contain the least information about nuisance factors. Most classification model $F$ can be divided into feature extractor $f$ with parameter $\theta_{f}$ and classifier $g$ with parameter $\theta_{g}$. In this way, the objective can be defined as:

\begin{equation}
    \min_{\theta_{f}} I(\psi;f(x;\theta_f))
\end{equation}\label{func:prob_form}
where input data $x \in X$, nuisance factor $\psi \in \Psi$, $I(a;b)$ denotes the mutual information between $a$ and $b$.

The intuition behind minimizing mutual information between $\psi$ and $f(x)$ is that the main task classifier $g$ takes extracted features $f(x)$ as input, so if $f(x)$ contains no information about nuisance factors, $g$ will not utilize nuisance factor features, thus being robust to nuisance factor perturbations.

\subsection{Nuisance-label Supervision (NLS) Module}\label{section:msm}
In this section, we present the formulation of the proposed nuisance-label supervision module. The module utilizes nuisance labels as extra supervision, in order to train a feature representation with the least information about nuisance factors.

\textbf{Overview}.
We give an overview of the whole network architecture in Fig.~\ref{fig:pipeline}. We utilize adversarial training to achieve the goal of "removing nuisance information" inspired by~\cite{ganin2014unsupervised}. We divide input data into two parts: 1) data w/o nuisance labels, 2) data w/ nuisance labels. On the one hand, both inputs are put through a feature extractor and classifier the same as normal action recognition pipeline. On the other hand, data with nuisance labels is additionally put into NLS module for training feature representation. In this process, nuisance labels are used as extra supervision signals and output an adversarial loss.

\textbf{Architecture}.
The NLS module consists of two parts: gradient reversal layer~\cite{ganin2014unsupervised} and nuisance factor classifier as is shown in Fig.~\ref{fig:msm}. The module input is extracted features and nuisance labels. The nuisance factor classifier is to predict labels of a certain nuisance factor, while corresponding nuisance labels are used for calculating a cross-entropy loss, which is widely used for classification tasks. The gradient reversal layer multiplies the gradient by a negative constant during the back-propagation process, making the input features as indistinguishable as possible for the nuisance factor classifier and ensuring the training is in an adversary way. 

\begin{equation}
    L_{\psi} = \mathbb{E}_{(x,\psi)\sim (X,\Psi)}L_{ce}(h(f(x;\theta_f);\theta_h), \psi)
\end{equation}
where $L_{ce}$ denotes cross-entropy loss.

Usually, there is more than one nuisance factor in a task, so we use multiple nuisance factor classifiers ($h_p$) to deal with each nuisance factor, then aggregate the outputs. The nuisance adversarial loss for multiple factors is defined as: $L_{\psi} = \sum_{p=1}^{N}L_{\psi^p}$, where superscript $p$ denotes the $p^{th}$ type of nuisance factors, and $N$ denotes the number of nuisance factors matter in the task. Fig.~\ref{fig:msm} also illustrates how we deal with multiple nuisance factors in NLS module.

\textbf{Training Procedure}.
Additional to removing nuisance factor effects, we certainly need the model to learn useful features for the main classification task:

\begin{equation}
    L_{cls} = \mathbb{E}_{(x,y)\sim (X,Y)}L_{ce}(g(f(x;\theta_f);\theta_g), y)
\end{equation}

So the final objective consists of classification loss and nuisance adversarial loss, which is formulated as follows:
\begin{equation}
    L = L_{cls} - \lambda L_{\psi}
\end{equation}
where, the parameter $\lambda$ controls the strength of the nuisance adversarial loss.

Based on the above objective function, the optimization problem is defined as follows:

\begin{equation}
\begin{aligned}
    L(\theta_{f}, \theta_{g}, \theta_{h}) &= L_{cls}(\theta_{f}, \theta_{g}) - \lambda L_{\psi}(\theta_{f}, \theta_{h}) \\
    (\hat\theta_{f}, \hat\theta_{g}) &= \arg \min L(\theta_{f}, \theta_{g}, \hat\theta_{h}) \\
    (\hat\theta_{h}) &= \arg \max L(\hat\theta_{f}, \hat\theta_{g}, \theta_{h})
\end{aligned}
\end{equation}


At training time, the nuisance factor classifier parameters $\theta_{h}$ minimize the nuisance adversarial loss and the action classifier parameters $\theta_{g}$ minimize classification loss as normal classification task. The feature extractor parameters $\theta_{f}$ \textit{minimize} classification loss while \textit{maximizing} nuisance adversarial loss simultaneously.

\subsection{Nuisance-label Generation}\label{section:sim_gen}
In this section, we formulate the generation of nuisance labels. As stated in Sec.~\ref{section:prob_stat}, we define data generation as $D(\phi, \psi)$. Usually, we only record labels for $\phi$ and throw out labels for $\psi$.

\begin{figure}[t]
\begin{center}
   \includegraphics[width=\linewidth]{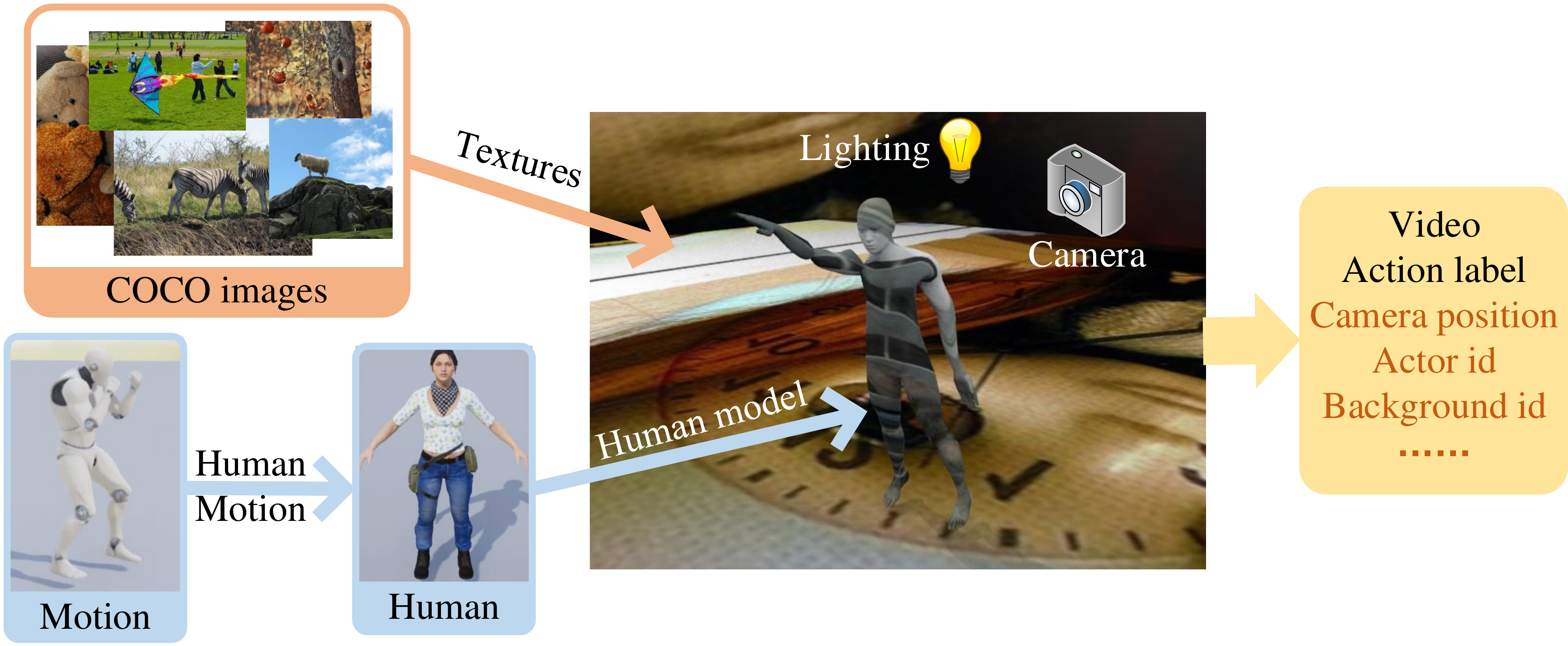}
\end{center}
    \vspace{-0.1in}
  \caption{Synthetic data is an effective way of getting nuisance labels and we build a pipeline for generating synthetic data. We follow the concept of domain randomization to set camera position, actor appearance and background in the virtual world. Together with action labels and video, we save these rendering parameters as nuisance labels.}
  \vspace{-0.1in}
\label{fig:sim_pipe}
\end{figure}

\begin{figure}[t]
\begin{center}
   \includegraphics[width=0.9\linewidth]{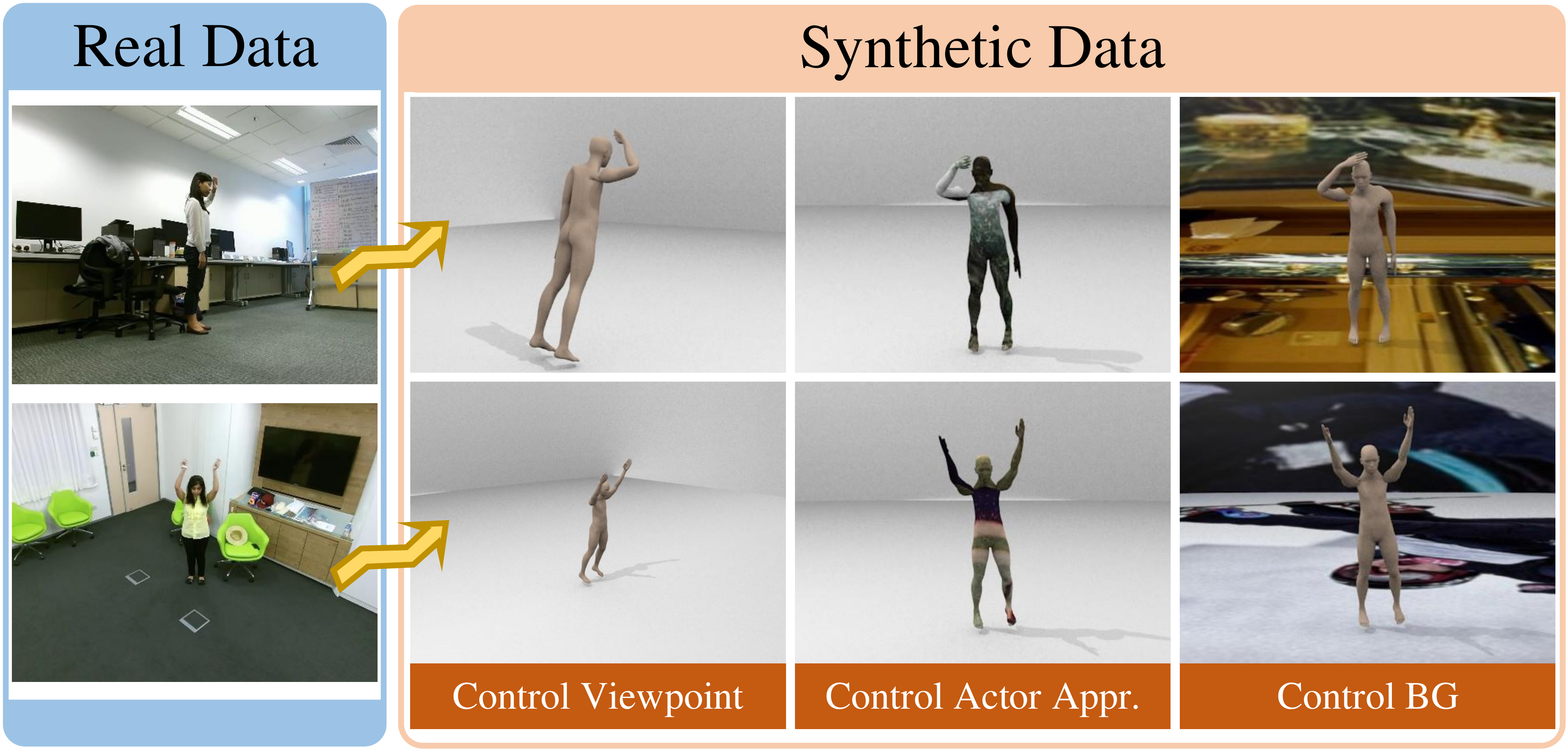}
\end{center}
    \vspace{-0.1in}
   \caption{Examples of synthetic data. The appearance is similar in each row and different in each column. But the action label of each row is the same. So a robust model needs to be invariant to these dramatic nuisance factors change. We utilize nuisance labels from synthetic data to achieve this goal. }
   \vspace{-0.1in}
\label{fig:sim_data}
\end{figure}

Data augmentation actually offers more information than people used to expect; the image processing parameters can be used in training as nuisance labels. For example, when generating images with Gaussian Noise, we can save the standard deviation values; when applying Median Blur, we can save the kernel size, these can both be used as nuisance labels.

Similarly, some existing real datasets also contain such information, i.e. metadata during data collection. People record metadata only for evaluation or showing randomization, while it can be used in training as well. For example, people collect NTU RGB-D dataset in a controlled way, that for each video, the recording environment, subject appearance, and camera viewpoint are all saved. This information can be regarded as nuisance labels.

Synthetic data generation offers a convenient way to manipulate multiple factors in the virtual world and we can easily reach the ground truth of these factors. In our work, we use CG renderer Blender\footnote{https://www.blender.org/} to manipulate viewpoint, textures in the virtual world, as is shown in Fig.~\ref{fig:sim_pipe} (more details are in Appendix). We save camera position, actor appearance id and background id as nuisance labels. The examples of our CG synthetic data are shown in Fig.~\ref{fig:sim_data}.

\subsection{Dependency Degree on Nuisance Factor}\label{section:define_dep_degree}
Besides serving as extra supervision signals, nuisance labels are also used to probe model dependency degree on certain factors. 

We design \textit{Dependency Degree on Nuisance Factor} to quantitatively measure how much a trained feature extractor focuses on nuisance factors. We attach a classifier to a trained feature extractor to predict the correct label of a certain nuisance factor. During training, weights in the feature extractor are frozen while parameters in the attached classifier are optimized. For a classification task, the worst performance corresponds to the random guess of inputs to classes, i.e. chance level performance; the chance level performance on a nuisance factor $\psi$ is denoted as $Acc_{Rand}(\psi)$. Thus we define the model dependency degree on a certain nuisance factor $\psi$ based on trained feature extractor $f$ as:

\begin{equation}
    Dep(\psi|f(x)) = \log{\frac{Prec(\psi|f(x))} {Acc_{Rand}(\psi)}\qquad}
\end{equation}
where $Prec(\psi|f(x))$ is the precision of the attached classifier. When the dependency degree goes down to zero, the nuisance factor classifier reaches chance-level performance, thus extracted features contain no information related to the nuisance factor. The definition is inspired by \textit{Dataset Bias} in~\cite{li2018resound} while it presents the dataset dependency and ours presents the model dependency on some factor.

\section{Experiments}
In this section, we explain setups and show results of our experiments. The setups including dataset and implementation details are shown in Sec.~\ref{section:dataset} and Sec.~\ref{section:imple_details}. The experiment results are organized by nuisance-label types, i.e. image processing parameters, real metadata and rendering parameters (Sec.~\ref{section:img_param}, ~\ref{section:real_metadata}, ~\ref{section:render_param}). And the ablation study of the relationship between model robustness and nuisance factor dependency is shown in Sec.~\ref{section:ablation}.

\subsection{Datasets}\label{section:dataset}
We evaluate our method on three datasets: MNIST-C~\cite{mu2019mnist}, UCF101~\cite{soomro2012ucf101} and NTU RGB-D~\cite{shahroudy2016ntu}.

\textbf{MNIST-C~\cite{mu2019mnist}}. This is a small-scale handwritten digits dataset, including a training set of 60,000 examples, and a testing set of 10,000 examples with 15 types of corruptions applied to the original MNIST test set.

\begin{figure}[t]
\begin{center}
   \includegraphics[width=\linewidth]{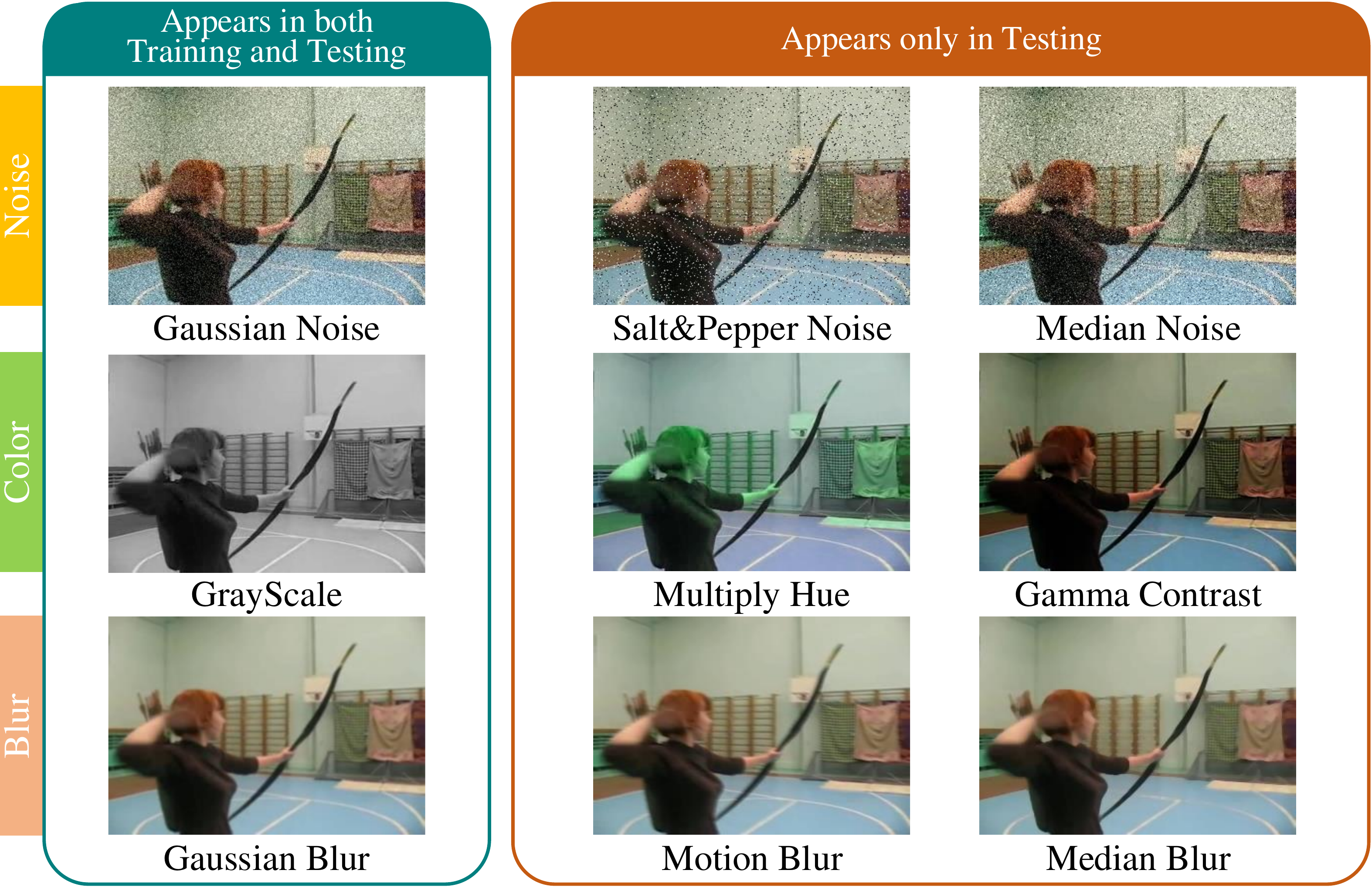}
\end{center}
    \vspace{-0.1in}
   \caption{Examples of image corruptions applied on UCF101 dataset. The left column is those used both in training augmentation and in testing. The right two columns are those used only for testing. We want to study model generalization to cases outside training augmentation.}
   \vspace{-0.1in}
\label{fig:ucf_corrup}
\end{figure}

\textbf{UCF101~\cite{soomro2012ucf101}}. \label{section:ucf101}
This is a large dataset with 101 human actions from videos in the wild; it consists of over 13k clips and 27 hours of video data with five types of action classes. We use it to prove our proposed method on 2D nuisance factors. We build a special test set to evaluate the model robustness on image corruptions, as shown in Fig.~\ref{fig:ucf_corrup}.

\textbf{NTU RGB-D~\cite{shahroudy2016ntu}}. \label{section:ntu}
This is a controlled human action dataset with 60 different action classes; it contains metadata besides action labels, including camera viewpoint, actor appearance, and recording environment. The metadata can be used as nuisance labels in NLS module. In addition, with these annotations, we create a more challenging data split, denoted as Cross-Nuisance (CN) split, where training and testing sets share no overlap on these three factors (details shown in Appendix). This split is for evaluating model robustness to 3D nuisance factor perturbations. 

\subsection{Implementation Details}\label{section:imple_details}
\textbf{Network architecture.}
For MNIST-C, we use the same architecture as in~\cite{madry2017towards}. For action recognition, we choose I3D~\cite{carreira2017quo} network as the baseline. We initialize the network with inflated weights from the 2D Inception network pretrained on ImageNet~\cite{deng2009imagenet}. At training time, we randomly sample 16 consecutive frames temporally, rescale the shorter side of all input frames to 226 pixels and randomly crop a $224 \times 224$ patch from each frame. The left-right flip is randomly applied as 2D data augmentation. At test time, we temporally sample three fragments evenly from the full video.

Our proposed NLS module is applied to the input feature map of the main task classifier. We choose $\alpha = 0.5$ for gradient reversal layer~\cite{ganin2014unsupervised}, which is the coefficient for reversed gradient during backpropagation. The nuisance classifier in NLS module is a three-layer MLP, with 1024 nodes in each layer. 

\textbf{Training Details.}
For MNIST-C, we train the network following the same rule as in~\cite{rusak2020increasing}. For action recognition, we set batch size as 8 on each GPU. We use SGD optimizer with momentum = 0.9, weight decay = 1e-4. The initial learning rate equals 0.01 and decays to 10\% when accuracy on validate set saturates. In order to prevent NLS from back-propagating noisy signals at the beginning of training, we set $\lambda$=0 for 10 epochs, then increase it to 0.05 for the rest of the training. We set $\lambda$ so small because we want $L_\psi$ to work as a regularization term, not to interfere with the main task training.

\textbf{Synthetic Data Generation.}
When generating synthetic data for NTU RGB-D, animation data captured from Microsoft Kinect~\cite{zhang2012microsoft} are used to drive 3D human skeletal meshes from MakeHuman\footnote{http://www.makehumancommunity.org/}. Following the concept of domain randomization~\cite{tobin2017domain}, the textures for human appearance and background are randomly sampled from 1000 images within the MSCOCO dataset~\cite{lin2014microsoft}; the camera is randomly set on 240 positions to record videos.

\subsection{NLS on Image Processing Parameters}\label{section:img_param}
In this section, we use corrupted data as augmentation and regard its processing parameters as nuisance labels. We regard GNT and ANT$1\times1$ in~\cite{rusak2020increasing} as the comparison, where Gaussian Noise and adversarial noise is added to part of the training data respectively as data augmentation.

\textbf{MNIST-C Experiments}.
We use experiments on MNIST-C~\cite{mu2019mnist} to evaluate models on image corruptions. MNIST-C is a simple but popular dataset for people to study image corruption. Our proposed method NLS module is applied to GNT, where we regard Gaussian Noise standard deviations as nuisance labels. The results are shown in Table~\ref{tab:mnist}. Our method surpasses GNT and ANT on both clean data and corrupted data.

\begin{table}
\begin{center}
\begin{tabular}{|l|c|c|}
\hline
Model & Clean & MNIST-C \\
\hline
Baseline & 99.13 & 86.86 \\
\hline
GNT & 99.40 & 92.39 \\
\hline
ANT1$\times$1 & 99.37 & 92.33 \\
\hline
GNT+NLS & \textbf{99.44} & \textbf{92.51} \\
\hline
\end{tabular}
\end{center}
    \vspace{-0.05in}
    \caption{\label{tab:mnist} Accuracy on MNIST (clean data) and MNIST-C (corrupted data) test sets. We compare our NLS with GNT and ANT1$\times$1, where models trained with noise and our NLS module.}
    \vspace{-0.05in}
\end{table}

\begin{table*}
\begin{center}
\begin{tabular}{|l||c||c|ccc|ccc|ccc|}
\hline
    \multirow{2}{*}{Model} & \multirow{2}{*}{Clean} & \multirow{2}{*}{Mean}  & \multicolumn{3}{c|}{Noise} & \multicolumn{3}{c|}{Color} & \multicolumn{3}{c|}{Blur}  \\
    \cline{4-12}
    &   &  & Gauss. & Salt. & Lap. & Gray & MulHue. & Contra. & Gauss. & Med. & Mot.  \\
\hline
Baseline & \textbf{93.41} & 77.85 & 63.17 & 45.15 & 47.65 & 90.58 & 90.57 & \textbf{92.62} & 92.53 & 87.80 & 90.60 \\
\hline
Aug & 92.30 & 90.18 & 90.36 & 87.17 & 89.07 & 91.58 & 89.69 & 91.43 & 92.00 & 89.61 & 90.70 \\
\hline
ANT1$\times$1 & 93.10 & 78.56 & 67.36 & 47.63 & 48.50 & 90.68 & \textbf{90.72} & 92.07 & 92.04 & 87.50 & 90.57 \\
\hline
Aug+NLS & 92.73 & \textbf{90.73} & \textbf{91.07} & \textbf{87.79} & \textbf{89.58} & \textbf{91.90} & 90.31 & 91.57 & \textbf{92.57} & \textbf{90.64} & \textbf{91.10} \\
\hline
\end{tabular}
\end{center}
    \vspace{-0.05in}
    \caption{\label{tab:ucf_nmm} Top-1 accuracy on UCF101 original and corrupted test sets. \textit{Aug} refers to data augmentation that three types of image corruptions are used as data augmentation during training.}
    \vspace{-0.05in}
\end{table*}

\textbf{UCF101 Experiments}.\label{section:ucf_nmm}
We further evaluate NLS module dealing with image corruptions on UCF101 dataset. We build a corrupted test set to evaluate the model robustness on image corruptions, as shown in Fig.~\ref{fig:ucf_corrup}. For each corruption, we study three types. As for the data augmentation baseline \textit{Aug}, we choose only one for augmentation during training. During testing, all the types are applied respectively. We add one nuisance factor classifier for each type, through which only augmented data with corresponding corruption will pass. For comparison, we train an ANT$1\times1$ and apply adversarial noise to each frame in a video sequence.

The results are shown in Table~\ref{tab:ucf_nmm}. We display accuracy on the clean test set and our designed corrupted test sets. The number is the mean value on three splits of UCF101.  \textit{Aug} has a good improvement on the corrupted test set especially on noise compared to the baseline but brings harm to clean data. Adding NLS further lifts model accuracy on clean data by 0.43\% and corrupted data by 0.55\% while adding adversarial noise to video data has marginal improvement. Though in some cases NLS is not the best, we highlight the consistent improvement compared with \textit{Aug}.

\subsection{NLS on Real Metadata}\label{section:real_metadata}
In this section, we evaluate NLS on real metadata in NTU RGB-D dataset, using its metadata, including camera viewpoint, actor appearance, and recording environment, as nuisance labels.

We conduct experiments on CN split (Sec.~\ref{section:ntu}) and study three factors: viewpoint, actor appearance, and background in action recognition tasks. We study both real metadata and rendering parameters on the same dataset NTU RGB-D, so we denote adding NLS on real metadata as \textit{Real NLS} and adding NLS on synthetic rendering parameters as \textit{Sim NLS} (shown in Sec.~\ref{section:render_param}). 

The results of Real NLS are shown in the top two rows in Table~\ref{tab:ntu_nmm}. Adding Real NLS brings an accuracy improvement of 1.07\% compared with the baseline model. The accuracy improvement is totally for free because no extra data is used. 

\textbf{Comparison with State-of-the-art}. We conduct experiments on NTU RGB-D standard CV and CS splits, and compare it with state-of-the-art methods using \textit{RGB only}. Note the data we use is totally the same as that in other methods. The results are shown in Table~\ref{tab:nmm_ntu_real}. Based on the strong baseline I3D, our proposed method reaches the best performance within RGB-based models, bringing accuracy improvement by 0.5\% and 0.4\% on two splits respectively. Besides, comparing with the results on CN split in Table~\ref{tab:ntu_nmm}, we observe that when the split is harder, the improvement brought by NLS is larger.

\begin{table}
\begin{center}
\begin{tabular}{|l|c|c|}
\hline
Model & Accuracy & F1 Score \\
\hline
Baseline & 64.50 & 66.92 \\
\hline
Real NLS & 65.57 & 67.32 \\
\hline
Sim Aug & 68.75 & 70.99 \\
\hline
Sim Aug+ANT$1\times1$ & 68.04 & 71.05 \\
\hline
Sim Aug+Sim NLS & 70.40 & 73.03 \\
\hline
Sim Aug+Real\&Sim NLS& \textbf{70.64} & \textbf{73.38} \\
\hline
\end{tabular}
\end{center}    
    \vspace{-0.05in}
    \caption{\label{tab:ntu_nmm} I3D accuracy and F1 score on NTU RGB-D Cross-Nuisance (CN) split. \textit{Real NLS} refers to applying NLS to real metadata and \textit{Sim NLS} refers to applying NLS to synthetic rendering parameters. \textit{Sim Aug} refers to using CG synthetic data as data augmentation.}
    \vspace{-0.05in}
\end{table}

\begin{table}
\begin{center}
\begin{tabular}{|l|c|c|c|}
\hline
Method & Modality & CS & CV   \\
\hline
Hands attention~\cite{Baradel17} & RGB+Skeleton & 84.8 & 90.6 \\
DA-Net~\cite{Wang_2018_ECCV}   & RGB+Flow & 88.1 & 92.0 \\
Pose evolution~\cite{Liu_2018_CVPR}& RGB+Depth & 91.7 & 95.3 \\
\hline
Hands attention~\cite{Baradel17}& RGB  & 75.6 & 80.5 \\
Pose evolution~\cite{Liu_2018_CVPR} & RGB  & 78.8 & 84.2 \\
Multi-task~\cite{Luvizon20182D3DPE}  & RGB  & 85.5 & - \\
Glimpse clouds~\cite{baradel18}  & RGB & 86.6 & 93.2 \\
I3D~\cite{carreira2017quo} & RGB & 90.2 & 95.2 \\
\hline
I3D + Real NLS& RGB & \textbf{90.7} & \textbf{95.6} \\
\hline
\end{tabular}
\end{center}
    \vspace{-0.05in}
    \caption{\label{tab:nmm_ntu_real} Comparison with state-of-the-art on standard splits of NTU RGB-D dataset. Our method achieves the highest performance within all the approaches using only RGB input.
    \vspace{-0.05in}
}
\end{table}

\subsection{NLS on Rendering Parameters}\label{section:render_param}
In this section, we evaluate NLS on rendering parameters, denoted as \textit{Sim NLS}. The real dataset setup is the same as in Sec.~\ref{section:real_metadata}. We generate CG synthetic data following rules in Sec.~\ref{section:sim_gen} as data augmentation, randomizing viewpoint, actor and background respectively, denoted as \textit{Sim Aug}. We add ANT1$\times$1 noise to synthetic training data as a comparison.

The results of Sim NLS are shown in Table~\ref{tab:ntu_nmm}. Adding CG synthetic data into training brings an accuracy improvement of 4.25\% while using Sim NLS further lifts the performance by 1.65\%. We notice ANT1$\times$1 has no improvement for 3D factor robustness, which shows the limitation of current 2D augmentation methods. We also study the effectiveness of adding Real and Sim NLS simultaneously, which reaches the best performance on CN test set.

\textbf{Domain Generalization}.\label{section:domain_gen}
We explore NLS module potential in domain generalization. The models are trained on \textit{pure} synthetic data and tested on real data. The synthetic data is generated following the rule of domain randomization~\cite{tobin2017domain}. 

The results of adding NLS on viewpoint, actor appearance, and background together are shown in Table~\ref{tab:domain_gen}. After adding NLS on synthetic rendering parameters, the model accuracy on real data improves observably by 2.24\%.

\begin{table}
\begin{center}
\begin{tabular}{|l|c|}
\hline
Model & Real Accuracy \\
\hline
DR & 18.16 \\
\hline
DR + NLS & \textbf{20.40} \\
\hline
\end{tabular}
\end{center}
    \vspace{-0.05in}
    \caption{\label{tab:domain_gen} Top-1 accuracy on NTU RGB-D real test set of I3D trained on domain randomization (DR) synthetic data w/ and w/o adding NLS. We train models purely on synthetic data to show that NLS has ability of reducing domain gap.}
    \vspace{-0.05in}
\end{table}

\subsection{Ablation Study}\label{section:ablation}

\subsubsection{NLS on single type of corruption}\label{section:ucf_single_nmm}
Besides applying NLS on multiple nuisance factors, we study the performance improvement on a single factor. We apply NLS to single image corruption on UCF101 dataset. We augment training data by one type of image augmentation and add the corresponding NLS.

The results in Table~\ref{tab:ucf_single_nmm} show that NLS on a single nuisance factor also improves model robustness. No matter NLS is applied to which factor, the performance is better than only using augmentation. One interesting phenomenon is that NLS brings an even larger lift to unseen corruptions, such as Salt and Pepper Noise and Median Blur, showing the ability for domain generalization.

\begin{table}
\begin{center}
\begin{tabular}{|l|c|c|c|}
\hline
Model & Gauss. & Salt. & Lap. \\
\hline
Baseline & 63.17 & 45.15 & 47.65\\
+Noise Aug & 90.28 & 87.70 & 89.24\\
+Noise Aug+NLS & \textbf{90.63} & \textbf{88.27} & \textbf{89.59}\\
\hline
 & Gray & MulHue. & Contra. \\
\hline
Baseline & 90.58 & 90.57 & 92.62 \\
+Color Aug & \textbf{92.57} & 91.00 & \textbf{92.14} \\
+Color Aug+NLS & \textbf{92.57} & \textbf{91.16} & 92.13 \\
\hline
 & Gauss. & Med. & Mot. \\
\hline
Baseline & 91.21 & 90.60 & 87.80 \\
+Blur Aug & 92.16 & 90.97 & 88.76 \\
+Blur Aug+NLS & \textbf{92.61} & \textbf{91.50} & \textbf{88.94} \\
\hline
\end{tabular}
\end{center}
    \vspace{-0.05in}
    \caption{\label{tab:ucf_single_nmm} Model accuracy on each corruption type in UCF101 test set. Only one type of corruption and its corresponding NLS are added to each model.}
    \vspace{-0.05in}
\end{table}

\subsection{Whether NLS helps reduce over-fitting}\label{section:reduce_overfit}
We design experiments to study whether NLS helps reduce over-fitting on the corresponding nuisance factor using NTU RGB-D dataset. In CN test set, all of the three factors are changed, so it is hard to say whether all the factors benefit from NLS. We apply NLS to one factor each time and evaluate it on three special test sets: \textit{View Diff}, \textit{BG Diff} and \textit{Actor Diff}. These sets have only \textbf{one} factor different from training data, e.g. \textit{View Diff} has the same actor and background settings but different viewpoint settings. We regard the performance improvement on \textit{Diff} test set as nuisance factor over-fitting reduction.

Table~\ref{tab:ntu_single_nmm} shows the results. After adding NLS, the corresponding factor over-fitting reduces obviously. Adding background NLS brings the smallest improvement, we explain it as the background in NTU RGB-D is less distinguishable than other factors. In addition, adding NLS on one nuisance factor also benefits robustness to other factors, e.g. accuracy on View Diff and BG Diff lifts sharply when adding NLS only on actor. While in most cases, adding NLS on three nuisance factors reaches the best performance.

\begin{table}
\begin{center}
\resizebox{\columnwidth}{!}{\begin{tabular}{|l|c|c|c|c|}
\hline
Model & View Diff & BG Diff & Actor Diff \\
\hline
Baseline & 70.96 & 87.59 & 87.67 \\
\hline
+Sim Aug & 73.58 & 88.46 & 88.06 \\
\hline
+Sim Aug+V NLS & 75.1 & 88.59 & 88.12 \\
\hline
+Sim Aug+B NLS & 74.43 & 88.89 & 87.71 \\
\hline
+Sim Aug+A NLS & 74.78 & 89.15 & \textbf{88.63} \\
\hline
+Sim Aug+3F NLS & \textbf{75.74} & \textbf{89.96} & 88.31 \\
\hline

\end{tabular}}
\end{center}
    \vspace{-0.05in}
    \caption{\label{tab:ntu_single_nmm} Top-1 accuracy on special test sets, the higher accuracy, the less over-fitting to a certain nuisance factor. \textit{View Diff} has only different viewpoint settings from training set. 3F refers to adding NLS on three nuisance factors and V, B, A refer to adding NLS on viewpoint, background, actor respectively.}
    \vspace{-0.05in}
\end{table}

\subsection{How much does NLS reduce model dependency on nuisance factors}
\begin{figure}[t]
\begin{center}
   \includegraphics[width=0.9\linewidth]{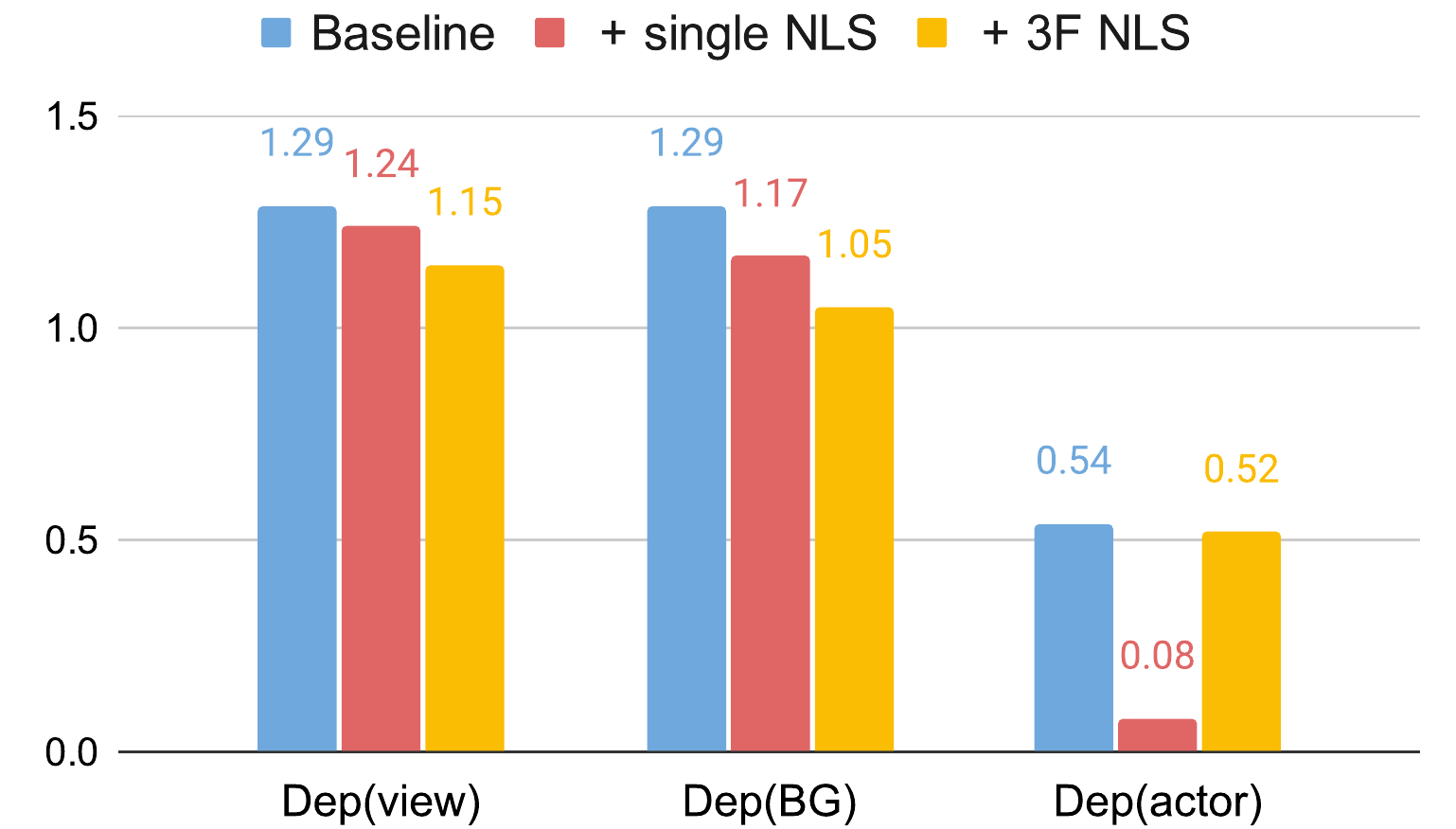}
\end{center}
    \vspace{-0.1in}
   \caption{Model dependency degree on three 3D nuisance factors w/ and w/o NLS. The random guess for view, BG and actor is 0.42\%, 0.1\% and 0.1\% respectively. }
   \vspace{-0.1in}
\label{fig:reduce_bias}
\end{figure}

We use nuisance labels to diagnose the model dependency degree on each 3D nuisance factor, further explaining the reason for robustness improvement. As defined in Sec.~\ref{section:define_dep_degree}, we train the attached nuisance factor classifier with feature extractor weights frozen. Fig.~\ref{fig:reduce_bias} shows the dependency degree changes, all the results are trained and tested on synthetic data. After adding NLS, the corresponding nuisance factor dependency degree decreases sharply. 

\section{Conclusion}
In this paper, we focus on the robustness problem caused by spurious correlation. We propose a novel module NLS using nuisance labels as extra supervision to break up spurious correlations, and we give three low-cost ways of acquiring nuisance labels, including image processing parameters, real metadata, and rendering parameters. Our proposed module effectively improves model robustness to nuisance factor perturbations. We demonstrate its effectiveness on MNIST-C, UCF101 and NTU RGB-D, handling image corruption and appearance changes, where our method all reaches the best performance. Furthermore, we utilize nuisance labels to diagnose the model dependency degree on a certain nuisance factor, explaining the reason why NLS helps reduce over-fitting. We hope that this work can contribute to the study of nuisance labels usage and enable more robust models.

{\small
\bibliographystyle{ieee_fullname}
\bibliography{egbib}
}

\end{document}